\documentclass{article}

\usepackage{arxiv}

\usepackage[utf8]{inputenc} 
\usepackage[T1]{fontenc}    
\usepackage{hyperref}       
\usepackage{url}            
\usepackage{booktabs}       
\usepackage{amsfonts}       
\usepackage{nicefrac}       
\usepackage{microtype}      
\usepackage{enumitem}       
\usepackage{graphicx}
\usepackage[square,numbers,sort&compress]{natbib}

\hypersetup{
    colorlinks=true,
    citecolor=black,
    linkcolor=blue,
    filecolor=magenta,      
    urlcolor=cyan,
    pdftitle={\texttt{sbi} reloaded: a toolkit for simulation-based inference workflows},
    }

\title{\texttt{sbi} reloaded: a toolkit for \\simulation-based inference workflows}

\author{
    Jan Boelts\thanks{Maintainer, equal contribution.}$\; ^{,\dag,1, 2, 3},\;$
    Michael Deistler$^{*,\dag,1, 2},\;$ \vspace{0.15cm} \\ \And
    Manuel Gloeckler\thanks{Core contributor.}$\; ^{\, ,\, 1, 2},\;$
    Álvaro Tejero-Cantero$^{\dag,\, 3,4},\;$
    Jan-Matthis Lueckmann$^{\dag,\, 5},\;$
    Guy Moss$^{\dag,\, 1, 2},\;$ \vspace{0.15cm} \\ \And
    Peter Steinbach\thanks{Major contributor.}$\; ^{\, ,\, 6},\;$
    Thomas Moreau$^{\ddag,\, 7},\;$
    Fabio Muratore$^{\ddag,\, 8},\;$
    Julia Linhart$^{\ddag,\, 7},\;$
    Conor Durkan$^{\ddag,\, 9},\;$ \vspace{0.15cm} \\ \And
    Julius~Vetter$^{1,2},\;$
    Benjamin~Kurt~Miller$^{10},\;$
    Maternus~Herold$^{3,11,12},\;$
    Abolfazl~Ziaeemehr$^{13},\;$ \\ \And
    Matthijs~Pals$^{1,2},\;$
    Theo~Gruner$^{14},\;$
    Sebastian~Bischoff$^{1,2,15},\;$
    Anastasia~N.~Krouglova$^{16,17},\;$
    Richard~Gao$^{1,2},\;$ \\ \And
    Janne~K.~Lappalainen$^{1,2},\;$
    Bálint~Mucsányi$^{1,2,18},\;$
    Felix~Pei$^{19},\;$
    Auguste~Schulz$^{1,2},\;$
    Zinovia~Stefanidi$^{1,2},\;$ \\ \And
    Pedro~L.~C.~Rodrigues$^{20},\;$
    Cornelius~Schröder$^{1,2},\;$
    Faried~Abu~Zaid$^{3},\;$
    Jonas~Beck$^{2,21},\;$
    Jaivardhan~Kapoor$^{1,2},\;$ \\ \And
    David~S.~Greenberg$^{22,23},\;$
    Pedro~J.~Gonçalves$^{17,24},\;$ \\ \And
    Jakob~H.~Macke$^{1,2,25}$ \vspace{0.3cm} \\
    jan.boelts@mailbox.org, michael.deistler@uni-tuebingen.de, jakob.macke@uni-tuebingen.de
}

\begin{document}
\maketitle

\vspace{-0.7cm} 
$^1$Machine Learning in Science, University of Tübingen;
$^2$Tübingen AI Center;
$^3$TransferLab, appliedAI Institute for Europe;
$^4$ML Colab, Cluster ML in Science, University of Tübingen;
$^5$Google Research;
$^6$Helmholtz-Zentrum Dresden-Rossendorf;
$^7$Université Paris-Saclay, INRIA, CEA, Palaiseau, France;
$^8$Robert Bosch GmbH;
$^{9}$School of Informatics, University of Edinburgh;
$^{10}$University of Amsterdam;
$^{11}$Research and Innovation Center, BMW Group;
$^{12}$Institute for Applied Mathematics and Scientific Computing, University of the Bundeswehr Munich, Germany;
$^{13}$Aix Marseille, INSERM, INS, France;
$^{14}$TU Darmstadt, hessian.AI, Germany;
$^{15}$University Hospital Tübingen and M3 Research Center;
$^{16}$Faculty of Science, B-3000, KU Leuven, Belgium;
$^{17}$VIB-Neuroelectronics Research Flanders (NERF) and imec, Belgium;
$^{18}$Methods of Machine Learning, University of Tübingen;
$^{19}$Neuroscience Institute, Carnegie Mellon University;
$^{20}$Université Grenoble Alpes, INRIA, CNRS, Grenoble INP, LJK, France;
$^{21}$Hertie Institute for AI in Brain Health, University of Tübingen;
$^{22}$Institute of Coastal Systems - Analysis and Modeling;
$^{23}$Helmholtz AI;
$^{24}$Departments of Computer Science Electrical Engineering, KU Leuven, Belgium;
$^{25}$Department Empirical Inference, Max Planck Institute for Intelligent Systems, Tübingen, Germany
\\[0.2cm]

\begin{abstract}
Scientists and engineers use simulators to model empirically observed phenomena. However, tuning the parameters of a simulator to ensure its outputs match observed data presents a significant challenge. Simulation-based inference (SBI) addresses this by enabling Bayesian inference for simulators, identifying parameters that match observed data and align with prior knowledge. Unlike traditional Bayesian inference, SBI only needs access to simulations from the model and does not require evaluations of the likelihood function. In addition, SBI algorithms do not require gradients through the simulator, allow for massive parallelization of simulations, and can perform inference for different observations without further simulations or training, thereby amortizing inference.
Over the past years, we have developed, maintained, and extended \texttt{sbi}, a PyTorch-based package\footnote{\texttt{sbi} is available at \href{https://github.com/sbi-dev/sbi}{\texttt{github.com/sbi-dev/sbi}} under the Apache 2.0 license.}
that implements Bayesian SBI algorithms based on neural networks. The \texttt{sbi} toolkit implements a wide range of inference methods, neural network architectures, sampling methods, and diagnostic tools. In addition, it provides well-tested default settings, but also offers flexibility to fully customize every step of the simulation-based inference workflow. Taken together, the \texttt{sbi} toolkit enables scientists and engineers to apply state-of-the-art SBI methods to black-box simulators, opening up new possibilities for aligning simulations with empirically observed data.
\end{abstract}

\section{Statement of need}
Bayesian inference is a principled approach for determining parameters consistent with empirical observations: Given a prior over parameters, a forward-model (defining the likelihood), and observations, it returns a posterior distribution. 
The posterior distribution captures the entire space of parameters that are compatible with the observations and the prior and it quantifies parameter uncertainty.
When the forward-model is given by a stochastic simulator, Bayesian inference can be challenging: (1) the forward-model can be slow to evaluate, making algorithms that rely on sequential evaluations of the likelihood (such as Markov-Chain Monte-Carlo, MCMC) impractical, (2) the simulator can be non-differentiable, prohibiting the use of gradient-based MCMC or variational inference (VI) methods, and (3) likelihood-evaluations can be intractable, meaning that we can only generate samples from the model, but not evaluate their likelihoods. 

Recently, simulation-based inference (SBI) algorithms based on neural networks have been developed to overcome these limitations \citep{papamakarios2016fast, papamakarios2019sequential, hermans2020likelihood}. Unlike classical methods from Approximate Bayesian Computation (ABC~\citep{sisson2018_chapter1}), these methods use neural networks to learn the relationship between parameters and simulation outputs. Neural SBI algorithms (1) allow for massive parallelization of simulations (in contrast to sequential evaluations in MCMC methods), (2) do not require gradients through the simulator, and (3) do not require evaluations of the likelihood, but only samples from the simulator. Finally, many of these algorithms allow for \emph{amortized} inference, that is, after a large upfront cost of simulating data for the training phase, they can return the posterior distribution for any observation without requiring further simulations or retraining.

To aid in the effective application of these algorithms to a wide range of problems, we developed the \texttt{sbi} toolkit. \texttt{sbi} implements a variety of state-of-the-art SBI algorithms, offering both high-level interfaces, extensive documentation and tutorials for practitioners, as well as low-level interfaces for experienced users and SBI researchers (giving full control over simulations, the training loop, and the sampling procedure). Since the original release of the \texttt{sbi} package \citep{tejerocantero2020sbi}, the community of contributors has expanded significantly, resulting in a large number of improvements that have made \texttt{sbi} more flexible, performant, and reliable. \texttt{sbi} now supports a wider range of amortized and sequential inference methods, neural network architectures (including normalizing flows, flow and score matching, and various embedding network architectures), samplers (including MCMC, variational inference, importance sampling, and rejection sampling), diagnostic tools, visualization tools, and a comprehensive set of tutorials on how to use these features.

The \texttt{sbi} package is already used extensively by the machine learning research community \citep{deistler2022truncated,gloecklervariational,muratore2022neural,gloeckler2023adversarial,dyer2022calibrating,wiqvist2021sequential,spurio2023bayesian,dirmeier2023simulation,gao2023generalized,gloeckler2024allinone,hermans2022crisis,linhart2024c2st,boelts2022flexible} but has also fostered the application of SBI in various research fields \citep{groschner2022biophysical,bondarenko2023embryo,confavreux2023meta,myers2024disinhibition,avecilla2022neural,lowet2023theta,bernaerts2023combined,mishra2022neural,dyer2022black,hashemi2023amortized,hahn2022accelerated,lemos2024field,deistler2022energy,rossler2023skewed,dingeldein2023simulation,jin2023bayesian,boelts2023simulation,gao2024deep,wang2024comprehensive}.

\section{Description}

\begin{figure}[h]
\includegraphics[width=\textwidth]{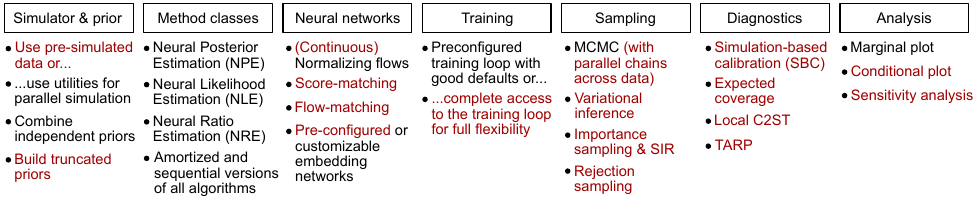}
\caption{{\bf Features of the \texttt{sbi} package.}
Components that were added since the initial release described in \citet{tejerocantero2020sbi} are marked in red.
}
\label{fig:fig1}
\end{figure}

\texttt{sbi} is a flexible and extensive toolkit for running simulation-based Bayesian inference workflows. \texttt{sbi} supports any kind of (offline) simulator and prior, a wide range of inference methods, neural networks, and samplers, as well as diagnostic methods and analysis tools (Fig.~\ref{fig:fig1}).

A significant challenge in making SBI algorithms accessible to a broader community lies in accommodating diverse and complex simulators, as well as varying degrees of flexibility in each step of the inference process. To address this, \texttt{sbi} provides pre-configured defaults for all inference methods, but also allows full customization of every step in the process (including simulation, training, sampling, diagnostics and analysis).

\textbf{Simulator \& prior:~} The \texttt{sbi} toolkit requires only simulation parameters and simulated data as input and no direct access to the simulator itself. However, if the simulator can be provided as a Python callable, \texttt{sbi} can optionally parallelize running the simulations from a given prior using Joblib \citep{joblib}. Additionally, \texttt{sbi} can automatically handle failed simulations or missing values, it supports both discrete and continuous parameters and observations (or mixtures thereof) and it provides utilities to flexibly define priors.

\textbf{Methods:~} \texttt{sbi} implements a wide range of neural network-based SBI algorithms, among them Neural Posterior Estimation (NPE) with various conditional estimators, Neural Likelihood Estimation (NLE), and Neural Ratio Estimation (NRE). 
Each of these methods can be run either in an \emph{amortized} mode, where the neural network is trained once on a set of pre-existing simulations results and then performs inference on \emph{any} observation without further simulations or retraining, or in a \emph{sequential} mode, where inference is focused on one observation to improve simulation efficiency with active learning.

\textbf{Neural networks and training:~} \texttt{sbi} implements a wide variety of state-of-the-art conditional density estimators for NPE and NLE, including normalizing flows \citep{papamakarios2021normalizing, greenberg2019automatic} (via nflows and Zuko \citep{nflows-repo, zuko-repo}), diffusion models \citep{song2021scorebased, geffner2023compositional, sharrock2022sequential}, mixture density networks~\citep{Bishop_94}, and flow matching \citep{lipman2023flow, dax2023flow} (via Zuko~\citep{zuko-repo}), as well as ensembles of any of these networks. \texttt{sbi} also implements a large set of embedding networks that can automatically learn summary statistics of (potentially) high-dimensional simulation outputs (including multilayer perceptrons, convolutional networks, and permutation-invariant networks). The neural networks can be trained with a preconfigured training loop with established default values, but \texttt{sbi} also allows full access over the training loop when desired.

\textbf{Sampling:~} For NLE and NRE, \texttt{sbi} implements a large range of samplers, including MCMC (with chains vectorized across observations), variational inference, rejection sampling, or importance sampling, as well as wrappers to use MCMC samplers from Pyro and PyMC \citep{bingham2019pyro, abril2023pymc}. \texttt{sbi} can perform inference for single observations or for multiple i.i.d.~observations, and can use importance sampling to correct for potential inaccuracies in the posterior if the likelihood is available.

\textbf{Diagnostics and analysis:~} The \texttt{sbi} toolkit also implements a large set of diagnostic tools, such as simulation-based calibration (SBC) \cite{talts2018validating}, expected coverage \cite{hermans2022crisis, deistler2022truncated}, local C2ST \citep{linhart2024c2st}, and TARP \citep{lemos2023sampling}. Additionally, \texttt{sbi} offers visualization tools, including marginal and conditional corner plots to visualize high-dimensional distributions, calibration plots, and wrappers for Arviz \citep{arviz_2019} diagnostic plots.

With \texttt{sbi}, our goal is to advance scientific discovery and computational engineering by making Bayesian inference accessible to a broad range of models, including those with inaccessible likelihoods, and to a broader range of users, including both machine learning researchers and domain practitioners. We have created an open architecture and embraced community-driven development practices to encourage collaboration with other machine learning researchers and applied scientists to join us in this long-term vision.

\section{Related software}

Simulation-based inference methods implemented in the \texttt{sbi} package require only access to simulated data, which can also be generated offline in other programming languages or frameworks. This sets \texttt{sbi} apart from toolboxes for traditional Bayesian inference, such as MCMC-based methods \citep{abril2023pymc, bingham2019pyro, gelman2015stan}, which rely on likelihood evaluations, and from probabilistic programming languages (e.g., Pyro \citep{bingham2019pyro}, NumPyro \citep{phan2019composable}, Stan \citep{gelman2015stan}, or Turing.jl \citep{ge2018t}), which typically require the simulator to be differentiable and implemented within their respective frameworks \citep{quera-bofarull2023}.

Since the original release of the \texttt{sbi} package, several other packages have emerged that implement neural network-based SBI algorithms. 
The Lampe~\citep{rozet_2021_lampe} package offers neural posterior and neural ratio estimation, primarily targeting SBI researchers with a low-level API and full flexibility over the training loop\footnote{The development of the Lampe package has stopped in favor of the \texttt{sbi} package in July 2024}.
The BayesFlow~\citep{bayesflow_2023_software} package focuses on a set of amortized SBI algorithms based on posterior and likelihood estimation. The Swyft~\citep{swyft} package specializes in algorithms based on neural ratio estimation. The sbijax~\citep{dirmeier2024simulationbasedinferencepythonpackage} package implements a set of inference methods in JAX.

\section{Author contributions}

This work represents a collaborative effort with contributions from a large and diverse team.  Author contributions are categorized as follows: Jan Boelts and Michael Deistler are the current maintainers and lead developers of the sbi package and contributed equally to this work.  Manuel Gloeckler, Álvaro Tejero-Cantero, Jan-Matthis Lueckmann, and Guy Moss have made substantial and sustained core contributions to the codebase and project direction. Peter Steinbach, Thomas Moreau, Fabio Muratore, Julia Linhart, and Conor Durkan have made major contributions to specific features or aspects of the package.  All other authors listed have contributed to the sbi package through code, documentation, or discussions. Jakob H. Macke provided overall project supervision and guidance.

\subsection*{Acknowledgements}
This work has been supported by
the German Federal Ministry of Education and Research (BMBF, projects `Simalesam', FKZ 01IS21055 A-B and `DeepHumanVision', FKZ:  031L0197B, and the Tübingen AI Center FKZ: 01IS18039A),
the German Research Foundation (DFG) through Germany’s Excellence Strategy (EXC-Number 2064/1, PN 390727645) and SFB1233 (PN 276693517), SFB 1089 (PN 227953431), SPP 2041 `Computational Connectomics', SPP 2298-2 (PN 543917411), the `Certification and Foundations of Safe Machine Learning Systems in Healthcare' project funded by the Carl Zeiss Foundation,
the Else Kröner Fresenius Stiftung (Project ClinbrAIn),
and the European Union (ERC, ``DeepCoMechTome'', ref. 101089288).
CD was supported by the EPSRC Centre for Doctoral Training in Data Science, funded by the UK Engineering and Physical Sciences Research Council (grant EP/L016427/1) and the University of Edinburgh.
BKM is part of the ELLIS PhD program, receiving travel support from the ELISE mobility program which has received funding from the European Union’s Horizon 2020 research and innovation programme under ELISE grant agreement No 951847.
DSG is supported by Helmholtz AI.
JL is a recipient of the Pierre-Aguilar Scholarship and thankful for the funding of the Capital Fund Management (CFM). ANK is supported by an FWO grant (G097022N).
TG was supported by “Third Wave of AI”, funded by the Excellence Program of the Hessian Ministry of Higher Education, Science, Research and Art.
TM and PLCR were supported from a national grant managed by the French National Research Agency (Agence Nationale de la Recherche) attributed to the ExaDoST project of the NumPEx PEPR program, under the reference ANR-22-EXNU-0004..
PS is supported by the Helmholtz Association Initiative and Networking Fund through the Helmholtz AI platform grant.
MD, MG, GM, JV, MP, SB, JKL, AS, ZS, JB are members of the International Max Planck Research School for Intelligent Systems (IMPRS-IS).

\bibliographystyle{unsrtnat}
\bibliography{paper}

\end{document}